\documentclass[letterpaper, 10 pt, conference]{ieeeconf}  

\IEEEoverridecommandlockouts                              

\usepackage{graphics} 
\usepackage{epsfig} 
\usepackage{mathptmx} 
\usepackage{times} 
\usepackage{amsmath} 
\usepackage{amssymb}  
\usepackage{url}
\usepackage{booktabs}
\usepackage{graphicx}
\usepackage{siunitx}

\DeclareMathOperator{\arctantwo}{arctan2}

\title{\LARGE \bf
Nearest-Neighbor-based Collision Avoidance for Quadrotors via Reinforcement Learning
}

\author{Ramzi Ourari*, Kai Cui*, Ahmed Elshamanhory and Heinz Koeppl
\thanks{*Authors contributed equally.}
\thanks{This work has been co-funded by the LOEWE initiative (Hesse, Germany) within the emergenCITY center and the Distr@l-project BlueSwarm (Project 71585164) of the Hessian Ministry of Digital Strategy and Development. The authors are with the Department of Electrical Engineering, Technische Universität Darmstadt, 64287 Darmstadt, Germany. Contact: {\tt\small  \{kai.cui, ahmed.elshamanhory, heinz.koeppl\}@bcs.tu-darmstadt.de}}%
}

\begin{document}

\maketitle
\thispagestyle{empty}
\pagestyle{empty}

\begin{abstract}
Collision avoidance algorithms are of central interest to many drone applications. In particular, decentralized approaches may be the key to enabling robust drone swarm solutions in cases where centralized communication becomes computationally prohibitive. In this work, we draw biological inspiration from flocks of starlings (Sturnus vulgaris) and apply the insight to end-to-end learned decentralized collision avoidance. More specifically, we propose a new, scalable observation model following a biomimetic nearest-neighbor information constraint that leads to fast learning and good collision avoidance behavior. By proposing a general reinforcement learning approach, we obtain an end-to-end learning-based approach to integrating collision avoidance with arbitrary tasks such as package collection and formation change. To validate the generality of this approach, we successfully apply our methodology through motion models of medium complexity, modeling momentum and nonetheless allowing direct application to real world quadrotors in conjunction with a standard PID controller. In contrast to prior works, we find that in our sufficiently rich motion model, nearest-neighbor information is indeed enough to learn effective collision avoidance behavior. Our learned policies are tested in simulation and subsequently transferred to real-world drones to validate their real-world applicability.

\end{abstract}


\section{INTRODUCTION}
The global market for unmanned aerial vehicles (UAVs) remains in the development stage with a potential market of \$45 billion in the civil sector alone \cite{kovalev2019analysis}, showing the enormous economic potential of drone swarms. Teams of robots and in particular quadcopters are found to be useful in many applications such as search-and-rescue missions \cite{karaca2018potential}, emergency communication \cite{camara2014cavalry} or package delivery \cite{shakhatreh2019unmanned}, the reason for which lies at least partially in their potential for low-level distributed processing applications such as decentralized object tracking and detection. At the same time, the usage of drone swarms in the real world adds various additional coordination challenges \cite{chmaj2015distributed}. As UAVs in swarms will often operate in close vicinity of each other, an important challenge for successful autonomous application of drone swarms is decentralized collision avoidance.

The study of collision avoidance has a long history and begins with traditional path-planning and sensor-based methods. Existing collision avoidance algorithms based on path-planning typically require expensive planning as well as full state information, complicating their usage in large drone swarms \cite{mellinger2012mixed}. On the other hand, although decentralized sensor-based methods can achieve very good performance -- e.g. ORCA \cite{alonso2013optimal} -- the disadvantage of such methods lies in their rigidity. While such algorithms may provide provable safety for the assumed model, they often lead to deadlock situations. In this context, end-to-end reinforcement learning (RL) may be able to provide better performance both in terms of deadlock occurrence and speed \cite{long2018towards}.

\begin{figure}
    \centering
    \vspace{0.2cm}
    \includegraphics[width=0.9\linewidth]{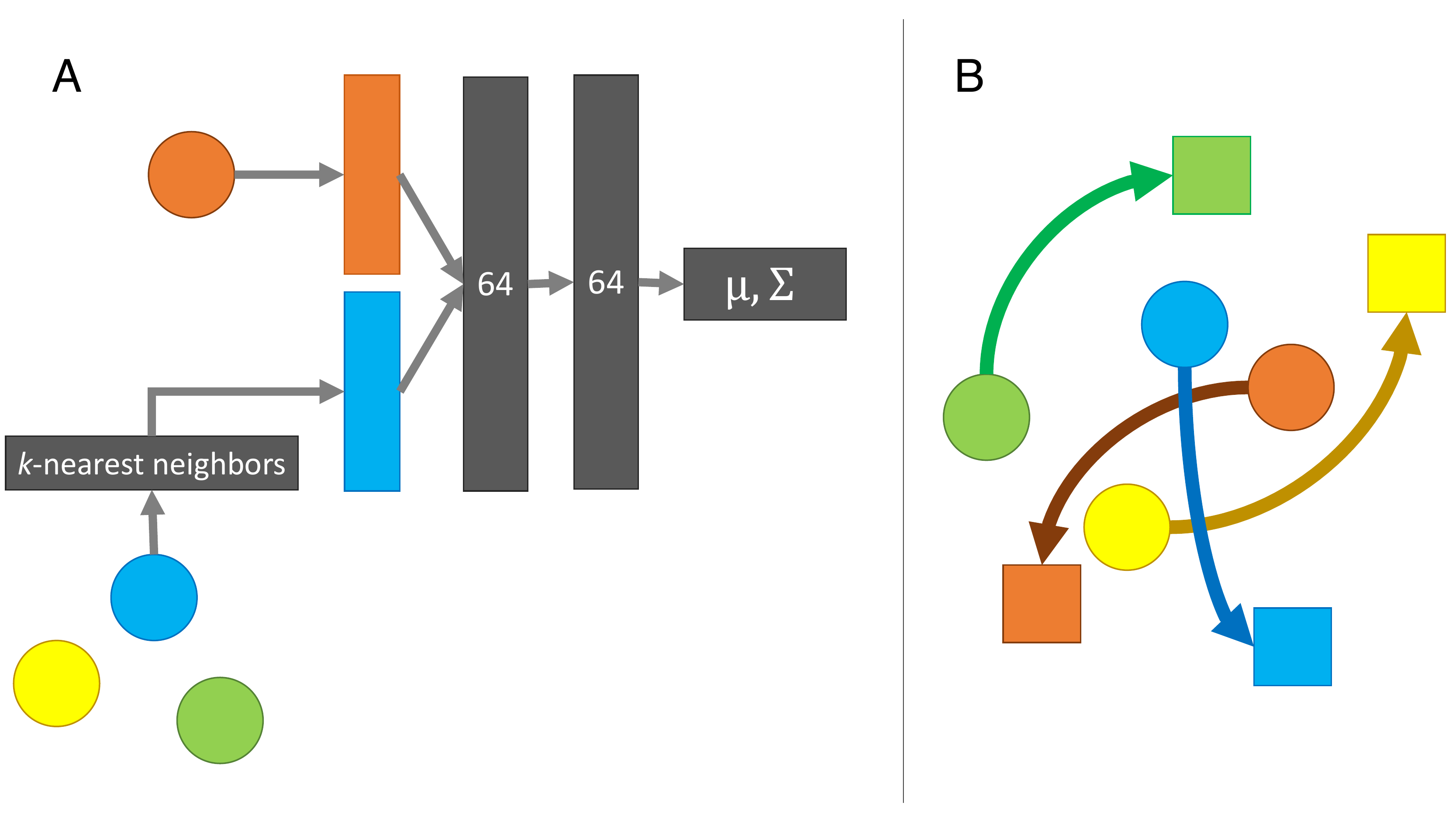}
    \caption{Illustration of our proposed architecture (BICARL). A: The agent's own observations (orange block) are concatenated with observations from the $k$ nearest neighbors (blue block, $k = 1$) and passed to a feed-forward neural network to obtain an immediate action distribution. B: The corresponding situation where agents need to negotiate their way toward their target while avoiding other dynamical agents. Each agent conditions their actions on observations of its $k$-nearest neighbors and their own state.} 
    \label{fig:hlvl}
\end{figure}

In this work, we propose a scalable, biologically well-motivated design to learning collision avoidance using end-to-end RL, i.e. {B}iologically-{I}nspired {C}ollision {A}voidance using {RL} (BICARL). In our method, we apply the biological insight into flocks of starlings (Sturnus vulgaris) interacting only with a limited number of their nearest neighbors to quadcopters in order to achieve decentralized collision avoidance that avoids crashes or getting stuck while learning successful policies in 30 minutes on two commodity desktop computers using an Intel Core i7-8700K CPU, 16 GB of RAM and no GPUs. This provides a scalable end-to-end learning approach to decentralized collision avoidance, reaching the performance of other state-of-the-art algorithms. Even though our motion model is more complex than models in prior works \cite{chen2017socially, everett2021collision}, we find that the added complexity allows improved collision avoidance performance and at the same time direct application to real world quadrotors by combining with conventional low-level controllers. As a result, we obtain a very practical deep RL approach to decentralized collision avoidance that remains scalable and applicable to arbitrary task specifications while requiring minimal information about the swarm, i.e. only from the nearest neighbor. Finally, we validate our algorithm both in simulation and real world application, verifying real world applicability.

\section{RELATED WORK}
Traditional methods for collision avoidance include planning approaches with full state information \cite{hamer2018fast}, or hand-engineered sensor-based approaches using velocity obstacles \cite{fiorini1998motion} and potential fields \cite{sigurd2003uav}, see also \cite{hoy2015algorithms} for a survey. 

Though the field of planning remains an active area of research \cite{honig2018trajectory}, typical disadvantages of planning-based approaches are the requirement of full state information and high computational cost, barring scalable application in decentralized drone swarms. On the other hand, sensor-based methods may be applied in decentralized swarms, but have the drawback of potential deadlocks. Amongst the sensor-based methods, Optimal reciprocal collision avoidance (ORCA) is a method based on velocity obstacles \cite{alonso2013optimal}, while force-based motion planning (FMP) is a state-of-the-art method based on potential fields \cite{semnani2019forcebased}.

\paragraph{Learning-based approaches}
 
Recently, there have been many attempts to use machine learning for collision avoidance \cite{long2018towards, willemsen2021mambpo}, typically outperforming traditional state-of-the-art methods in terms of success rate and speed by learning a collision avoidance rule via reinforcement learning \cite{chen2017decentralized}, though many prior works focus on simplified dynamics models and consider application only to small swarms or single drones \cite{kahn2017uncertainty}. 

The GA3C-CADRL algorithm \cite{everett2021collision} applies LSTMs in a single integrator model to learn a reaction to the position and velocities of all other agents in order to jointly avoid collisions. From the imitation learning realm, GLAS \cite{riviere2020glas} focuses on imitating decisions of a global privileged planner with safety module. \cite{semnani2020multi} extend GA3C-CADRL to 3D dynamics and fuses FMP \cite{semnani2019forcebased} with RL in a hybrid algorithm. Finally, \cite{wang2020two} applies imitation learning on ORCA for initialization, and then refines using RL. All of the aforementioned approaches except for \cite{chen2017socially} remain conditioned on information of all agent's (relative) positions and velocities. Similar to our work, \cite{chen2017socially} proposes RL-based collision avoidance via nearest neighbor information in the single integrator model, though their approach remains limited to very small swarms and in \cite{everett2021collision} was found to become stuck in large swarms. 

\paragraph{Biological inspiration}

In the field of robot navigation, there exist a great number of works inspired by biology. To name a few, one can find navigation algorithms inspired by bugs \cite{mcguire2019comparative}, optical flow navigation inspired by honey bees \cite{green2008optic} or rule-based swarming models inspired by collective motion \cite{vicsek2012collective}, see also \cite{hoy2015algorithms} for a review.

In our work, we give a biological motivation for the nearest neighbor method. To be precise, we take inspiration from the behavior of flocks of starlings following a topological nearest-neighbor interaction rule in order to achieve robust swarm flight behavior \cite{young2013starling}. In prior works, this type of biological insight has inspired flocking control design \cite{liu2020leader} and real-world deployment of drone swarms \cite{petravcek2020bio}. Somewhat more related, \cite{zhu2020multi} implements end-to-end-learned flocking control based on classical swarming models. However, their focus remains on flocking, and their observation model is fully observed and therefore not decentralized.

\section{MODEL}
In this work we consider partially observable stochastic games (POSG) as the typical setting of multi-agent reinforcement learning. Formally, in our setting a POSG is a tuple $(I, X, U, Z, T, r, p, \gamma)$. The index set $I = \{1, \ldots, N\}$ is the set of agents, $X$ is the state space, $U = \bigotimes_{i \in I} U_i$ is the joint action space and $Z = \bigotimes_{i \in I} Z_i$ is the joint observation space, where $U_i$ and $Z_i$ denote the action and observation spaces of each agent respectively. In our work, we use a deterministic transition function $T \colon X \times U \to X$ and a random observation emission density $p \colon Z \times X \to \mathbb R_{\geq 0}$. The reward function $r \colon X \times U \to \mathbb R^N$ and the discount factor $\gamma \in (0,1)$ give rise to the maximization objective 
\begin{align}
    J_i(\boldsymbol \pi) = \mathbb E_{\boldsymbol \pi} \left[ \sum_{t=0}^\infty \gamma^t r_i(x_t, u_t) \mid x_0 \sim \mu_0 \right]
\end{align}
of agent $i$ with initial state distribution $\mu_0$ over joint policies $\boldsymbol \pi = (\pi^1, \ldots, \pi^N)$ with $\pi^i \colon U \times Z_i \to \mathbb R_{\geq 0}$ and
\begin{align}
    z_t \sim p(\cdot \mid x_t), \quad u_t^i \sim \pi^i(\cdot \mid z_t^i), \quad x_{t+1} = T(x_t, u_t)
\end{align} 
for $t \geq 0, i \in I$ and $u_t \equiv (u_t^1, \ldots, u_t^N), z_t \equiv (z_t^1, \ldots, z_t^N)$.

\subsection{Dynamics}
In our work, we will consider both a 2D and a 3D case. The perhaps simplest model studied in multi-agent collision avoidance is the 2D single integrator model used in most prior works such as \cite{alonso2013optimal, everett2021collision}, where the states $x_t \equiv (\mathbf p^i_t, \mathbf p^{i,*}_t)_{i \in I}$ consist of the $\mathbb R^2$-valued position $\mathbf p^i_t = (x^i_t, y^i_t)$ as well as goal position $\mathbf p^{i,*}_t = (x^{i,*}_t, y^{i,*}_t)$ of agents $i$. As actions $u_t^i \equiv \mathbf v^i_t$, each agent may choose their $\mathbb R^2$-valued velocity $\mathbf v^i_t$ directly as an action under the constraint $\lVert \mathbf v^i_t \rVert_2 \leq v_{\mathrm{max}}$, leading to the deterministic transition function defined by
\begin{align}
    \mathbf p^i_{t+1} = \mathbf p^i_t + \Delta t \cdot \mathbf v^i_t
\end{align}
for time step size $\Delta t \geq 0$. 

We propose a more complex model with some degree of momentum by using the following modified double integrator model, where the state $x_t \equiv (\mathbf p^i_t, \mathbf v^i_t, \theta^i_t, \omega^i_t, \mathbf p^{i,*}_t)_{i \in I}$ is given not only by the positions, but also by the $\mathbb R^2$-valued velocities $\mathbf v^i_t$ as well as the $\mathbb R$-valued yaw angle and its associated angular rate $\theta^i_t, \omega^i_t$. 

Note that this model can alternatively be understood as part of our algorithm on top of the single integrator model, i.e. the algorithm keeps track of any additional states and simulates the modified double integrator dynamics to choose velocities at every time step. Therefore, results for this more complex dynamics model are nonetheless applicable and comparable to the single integrator case.

An action $u_t^i \equiv (\tilde {\mathbf v}^i_t, \tilde {\omega}^i_t)$ of agent $i$ with chosen target velocity $\tilde {\mathbf v}^i_t \in \mathbb R^2$ and target angular velocity $\tilde {\omega}^i_t \in \mathbb R$ leads to the deterministic update
\begin{align}
    \mathbf p^i_{t+1} &= \mathbf p^i_t + \mathbf v^i_t \cdot \Delta t  , \\
    \mathbf v^i_{t+1} &= \mathbf v^i_t - c_v (\mathbf R(\theta^i_t) \cdot \tilde {\mathbf v}^i_t - \mathbf v^i_t) \cdot \Delta t , \\
    \theta^i_{t+1} &= \theta^i_t + \omega^i_t \cdot \Delta t , \\
    \omega^i_{t+1} &= \omega^i_t - c_\omega (\tilde {\omega}^i_t - \omega^i_t) \cdot \Delta t
\end{align}
with $\lVert \tilde {\mathbf v}^i_t \rVert_2 \leq v_{\mathrm{max}}$, $\left| \tilde {\omega}^i_t \right| \leq \omega_{\mathrm{max}}$ and yaw rotation matrix
\begin{align}
    \mathbf R(\theta^i_t) = 
    \begin{bmatrix}
        \cos \theta^i_t & -\sin \theta^i_t \\
        \sin \theta^i_t & \cos \theta^i_t
    \end{bmatrix} \, .
\end{align} 

Importantly, although the yaw angle is not strictly required, we add the yaw angle to empirically obtain significantly better performance, as the added model richness allows agents to implicitly save information without requiring e.g. recurrent policy architectures. This model of medium complexity will at the same time allow us to directly use the desired velocity output as a set point of more traditional low-level quadrotor controllers such as PID controllers. 

Note that depending on the specific task, we add task-specific transitions for the goal position, see the section on experiments. For the 3D case, we simply add another $z$-coordinate to position and velocity that remains unaffected by the rotation matrix. It should be further noted that we observe no discernible difference when applying some small amount of noise to the dynamics.

\subsection{Observation model}
We let the observations of an agent $i$ be randomly given by the own position and velocity as well as the relative bearing, distance and velocity of other agents inside of the sensor range $K > 0$ and the goal, i.e.
\begin{multline}
    z_t^i \equiv (\hat{\mathbf p}^i_t, \hat{\mathbf v}^i_t, d^{i,*}_t, \phi^{i,*}_t, \\
    \{ (d^{i,j}_t, \phi^{i,j}_t, \mathbf v^{i,j}_t) \mid j \neq i \colon \lVert \mathbf p^j_t - \mathbf p^i_t \rVert_2 \leq K \})
\end{multline}
where the observations are Gaussian distributed according to
\begin{align}
    \hat{\mathbf p}^i_t &\sim \mathcal N(\mathbf p^i_t, \sigma_p^2), \quad
    \hat{\mathbf v}^i_t \sim \mathcal N(\mathbf v^i_t, \sigma_v^2), \\
    d^{i,*}_t &\sim \mathcal N(\lVert \mathbf p^{i,*}_t - \mathbf p^i_t \rVert_2, \sigma_d^2), \\
    \phi^{i,*}_t &\sim \mathcal N(\varphi^{i,*}_t \sigma_\phi^2), \\
    d^{i,j}_t &\sim \mathcal N(\lVert \mathbf p^j_t - \mathbf p^i_t \rVert_2, \sigma_d^2), \\
    \phi^{i,j}_t &\sim \mathcal N(\varphi^{i,j}_t, \sigma_\phi^2), \\
    \mathbf v^{i,j}_t &\sim \mathcal N(\mathbf v^{j}_t - \mathbf v^{i}_t, \sigma_v^2)
\end{align}
with noise standard deviations $\sigma_p, \sigma_v, \sigma_d, \sigma_\phi > 0$ and bearing angles $\varphi^{i,*}_t = \arctantwo(y^{i,*}_t - y^i_t, x^{i,*}_t - x^i_t)$, $\varphi^{i,j}_t = \arctantwo(y^j_t - y^i_t, x^j_t - x^i_t)$ where $\arctantwo(y,x)$ is defined as the angle between the positive $x$-axis and the ray from $0$ to $(x,y)$. Note further that the observations allow application of e.g. ORCA \cite{alonso2013optimal} and FMP \cite{semnani2019forcebased}. In the 3D case, we additionally observe the $z$-coordinate difference to the target and other agents with Gaussian noise of standard deviation $\sigma_p$.

\subsection{Reward function}
As the reward function for all of our experiments, for each agent $i \in I$ we define a term for reaching the goal position and for avoiding collisions each by
\begin{multline}
    r_i(x_t, u_t) = c_p \langle \mathbf v^i_t, \mathbf p^{i,*}_t - \mathbf p^i_t \rangle \\
    - c_c \sum_{j \in I \setminus \{ i \}} \mathbf 1 \left( \lVert \mathbf p^j_t - \mathbf p^i_t \rVert_2 \leq C \right)
\end{multline}
with desired target position $\mathbf p^{i,*}_t \in \mathbb R^2$, avoidance radius $C \geq 0$ and reward / collision penalty coefficients $c_p, c_c \geq 0$, where $\langle \cdot, \cdot \rangle$ denotes the standard dot product. 

Note that this objective is selfish, i.e. agents are only interested in their own collisions and reaching their own goal. This manual choice of reward function alleviates the well-known multi-agent credit assignment problem in multi-agent RL \cite{hernandez2019survey}, as shared, cooperative swarm reward functions are difficult to learn due to the increasing noise from other agent's behaviors as the number of agents rises.

\section{METHODOLOGY}
In this work, we propose a biologically-inspired design principle to learning collision avoidance algorithms. Recently, it was found that swarms of starlings observe only up to seven neighboring birds to obtain robust swarm flight behavior \cite{young2013starling}. This implies that it should be sufficient to use a similar observation reduction for collision avoidance. Further, it is well-known that the multi-agent RL domain suffers from the combinatorial nature of multi-agent problems \cite{hernandez2019survey}. Hence, the reduction of observations to e.g. the closest $k$ agents can greatly help learning effective behavior.

It is known that tractable exact solution methods and theoretical guarantees in the context of POSGs are sparse even in the fully cooperative case \cite{zhang2021multi}. Instead, we apply independent learning, i.e. we ignore the non-stationarity of other agents and solve assumed separate single-agent problems via RL \cite{tan1993multi}. Furthermore, we share a policy between all agents via parameter sharing \cite{gupta2017cooperative} and use the PPO algorithm to learn a single, shared policy \cite{schulman2017proximal}. For faster learning, we collect rollout trajectories from our simulation in parallel on two machines and use the RLlib \cite{liang2018rllib} implementation of PPO. We implemented our environment in Unity ML-Agents \cite{juliani2018unity}. 

\begin{table}
\centering
\caption{Hyperparameters and parameters used in all experiments.}
\label{table:hyperparameters}
\begin{tabular}{@{}ccc@{}}
\toprule
Symbol     & Function          & Value     \\ \midrule
$\Delta t$          &    Time step size   & \SI{0.02}{\second}    \\
$c_v$          &    Velocity coefficient   & \SI{1}{\per \second}    \\
$c_w$          &    Angular rate coefficient   & \SI{1}{\per \second}     \\
$\sigma_p, \sigma_v$ &   Noise standard deviations & \SI{1}{\milli\metre}, \SI{10}{\milli\metre}\\
$\sigma_d, \sigma_\phi$ &   Noise standard deviations & \SI{1}{\milli\metre}, \SI{0.1}{\milli\metre}\\
$C$          &    Avoidance radius   & \SI{7}{\metre}     \\
$c_p$          &    Reward coefficient   & \SI{0.3}{\second \per \square \metre}     \\
$c_c$         &    Penalty coefficient   & $1$     \\
$v_{\mathrm{max}}$      & Maximum velocity         & \SI{30}{\metre \per \second} \\
$\omega_{\mathrm{max}}$  & Maximum angular velocity   & \SI{15}{\per \second} \\
$K$                     & Sensor range  & \SI{10}{\metre} \\
$k$                     & Number of considered neighbors  & 1 \\
$\gamma$ &   Discount factor &  $0.99$\\
\\ \midrule
        &  PPO                  & \\ \midrule
$l_{r}$ &   Learning rate & \num{5e-5}\\
$\beta$ &   KL coefficient & $0.2$ \\
$\epsilon$ &  Clip parameter & $0.3$ \\
$B$ &  Training batch size &  $50000$ \\
$B_{m}$ & SGD Mini batch size &  $2500$ \\
$M$ &  SGD iterations & $20$ \\ \bottomrule
\end{tabular}
\end{table}

We parametrize our policy by a simple feedforward network consisting of two hidden layers with 64 nodes, and ReLU activations except for the output layer which outputs parameters of a diagonal Gaussian distribution over actions. Actions are clipped after sampling to fulfill constraints. We preprocess observations by reducing to the information of the nearest $k$ neighbors in $L_2$ norm on the positions $\mathbf p^i_t$. All results in this work are for $k=1$, i.e. we use observations $(d^{i,j}_t, \phi^{i,j}_t, \mathbf v^{i,j}_t)$ of the nearest neighbor $j$. Crucially, this observation reduction works well in our experiments, and it allows us to show that information of neighboring agents limited to the nearest neighbor information is indeed sufficient for collision avoidance. A resulting advantage of our design is that the resulting policy itself is very simple and memoryless.

Note that our policy architecture is very small and thus allows sufficiently fast computation on low-powered microcontrollers and UAVs such as the Crazyflie \cite{giernacki2017crazyflie} used in our real world experiments. In comparison, previous works use Deep Sets \cite{zaheer2017deep} or LSTMs \cite{everett2021collision} to process a variable number of other agents, which will scale worse in large swarms and is more computationally costly to learn and apply.

The scenarios considered in this work include formation change and package delivery. In the formation change scenario, all drones start on a circle and have their goal set to the opposite side. In the package delivery scenario, drones are continuously assigned new random package locations and destinations after picking and dropping off packages, where the locations and destinations are uniformly distributed e.g. on a line, circle or in a rectangle.

\section{EXPERIMENTS AND RESULTS}
\begin{table} 
\centering
\caption{Performance of BICARL (ours), ORCA \cite{alonso2013optimal}, and FMP \cite{semnani2019forcebased} in the package delivery task, averaged over 4 runs of 100 seconds.}
\label{table:circle-comparison}
\begin{tabular}{@{}ccccccc@{}}
\toprule
Test setup & \multicolumn{3}{c}{Average collected packages} \\ \midrule
\# agents  & BICARL          & ORCA         & FMP   \\ \midrule
4          &    \textbf{15.25}   & 1.5      & 3     \\
6          &    \textbf{8.66}    & 1.16     & 2     \\
8          &    \textbf{7.62}    & 1.25     & 2.12  \\
10         &    \textbf{5.8}     & 0        & 1.2   \\
12         &    \textbf{4.41}    & 0        & 0.41  \\
14         &    \textbf{4.14}    & 0        & 0.25  \\ \bottomrule
\end{tabular}
\end{table}

\begin{figure}
	\centering
	\includegraphics[width=0.55\linewidth]{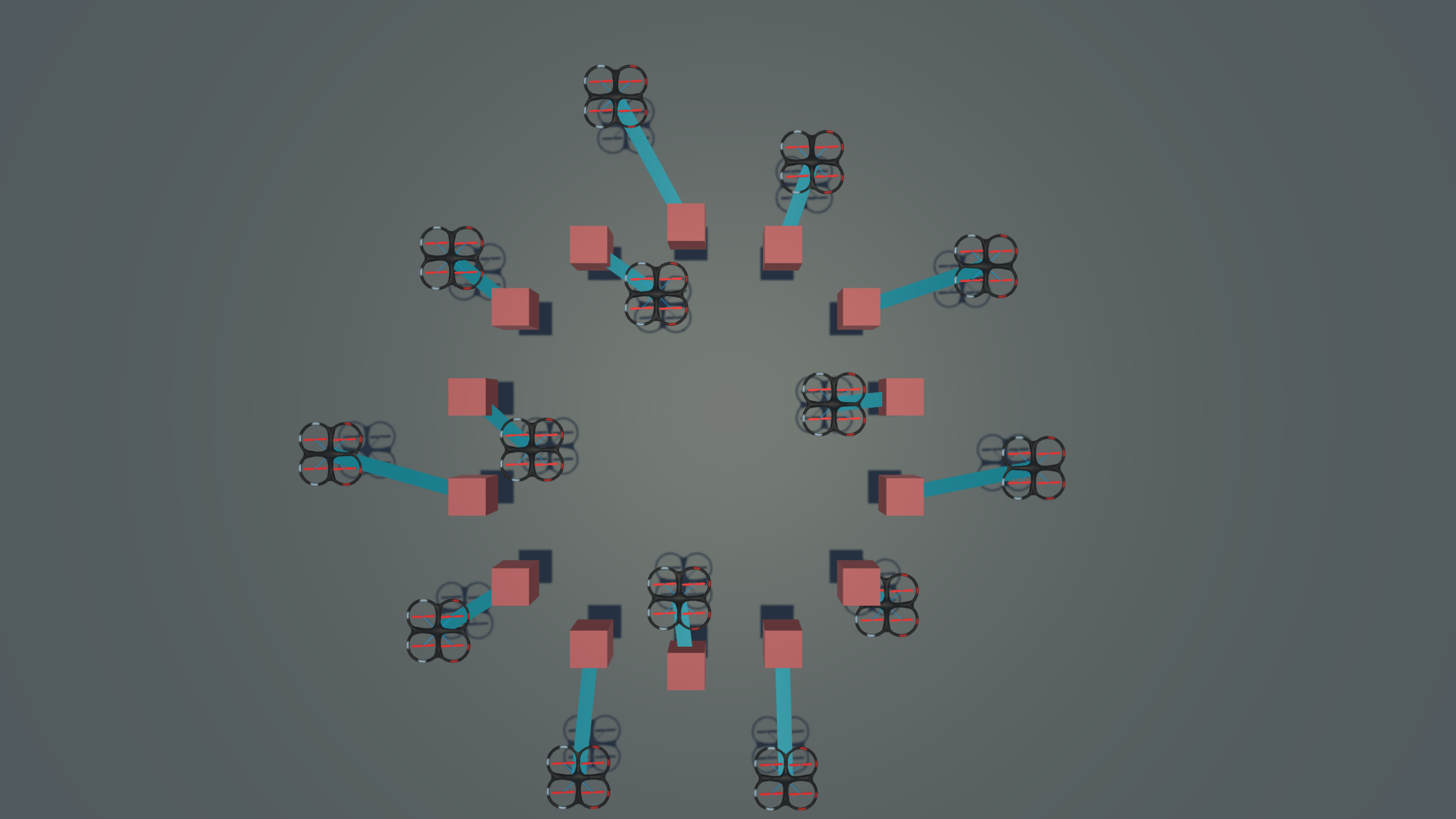}
	\caption{A constellation where FMP is stuck in package delivery: At this number of agents, the computed forces nullify each other, resulting in freezing agents. The red squares visualize the current goal of agents and the blue lines signify which agent is assigned which goal.}
	\label{fig:fmp_stuck_circle}
\end{figure}

\begin{table*} 
\centering
\caption{Performance analysis of BICARL (ours), ORCA \cite{alonso2013optimal}, and FMP \cite{semnani2019forcebased} in circle formation change, averaged over 5 runs.}
\label{table:comparison}
\renewcommand{\arraystretch}{1.21}
\begin{tabular}{@{}cllccccccccc@{}}
\toprule
\multicolumn{3}{c}{Test setup} &
  \multicolumn{3}{c}{Extra time to goal (\si{\second})} &
  \multicolumn{3}{c}{Minimal distance to other agents (\si{\metre})} &
  \multicolumn{3}{c}{Extra travelled distance (\si{\metre})} \\ \midrule
\multicolumn{3}{c}{\# agents} &
  BICARL &
  ORCA &
  FMP &
  BICARL &
  ORCA &
  FMP &
  BICARL &
  ORCA &
  FMP \\ \midrule
\multicolumn{3}{c}{5}  & 4.02 & 14.52 & \textbf{1.86}  & \textbf{12.38} & 4.94 & 6.18 & 1.28 & 1.17  & \textbf{1.11}  \\
\multicolumn{3}{c}{10} & 4.51 & 15.31 & \textbf{1.95}  & \textbf{7.01}  & 4.78 & 3.78 & 1.26 & 1.65 & \textbf{1.15} \\
\multicolumn{3}{c}{15} & 4.87 & 14.98 & \textbf{2.15} & \textbf{6.14}  & 3.42 & 4.44 & 1.27 & 1.82 & \textbf{1.19} \\
\multicolumn{3}{c}{20} & 6.51 & 18.68 & \textbf{2.34} & 3.14  & \textbf{4.74} & 3.87 & 1.45 & 2.7  & \textbf{1.26} \\
\multicolumn{3}{c}{25} & 7.52 & 20.13 & \textbf{2.28} & 3.3   & \textbf{4.7} & 3.34 & 1,60 & 3.56 &  \textbf{1.23}\\
\multicolumn{3}{c}{30} & 7.42 & 31.51 & \textbf{3.66} & 3.52  & \textbf{4.7} & 2.94 & 1,61 & 4.68  & \textbf{1.38} \\
\multicolumn{3}{c}{35} & \textbf{7.81} & 41.15 & N/A & 2.52  & \textbf{4.72} & N/A & \textbf{1.60} & 5.71 & N/A \\
\multicolumn{3}{c}{40} & \textbf{9.18} & 45.64 & N/A & 2.35  & \textbf{2.75} & N/A & \textbf{1.79} & 10.17 & N/A \\
\multicolumn{3}{c}{45} & \textbf{8.91} & 76.25 & N/A & 1.94  & \textbf{3.21} & N/A & \textbf{1.76} & 8.14 & N/A \\
\multicolumn{3}{c}{50} & \textbf{10.1} & 81.03 & N/A & 1.67  & \textbf{2.66} & N/A & \textbf{1.87} & 17.95 & N/A \\ \bottomrule
\end{tabular}
\end{table*}

\begin{figure}
	\centering
	\includegraphics[width=0.95\linewidth]{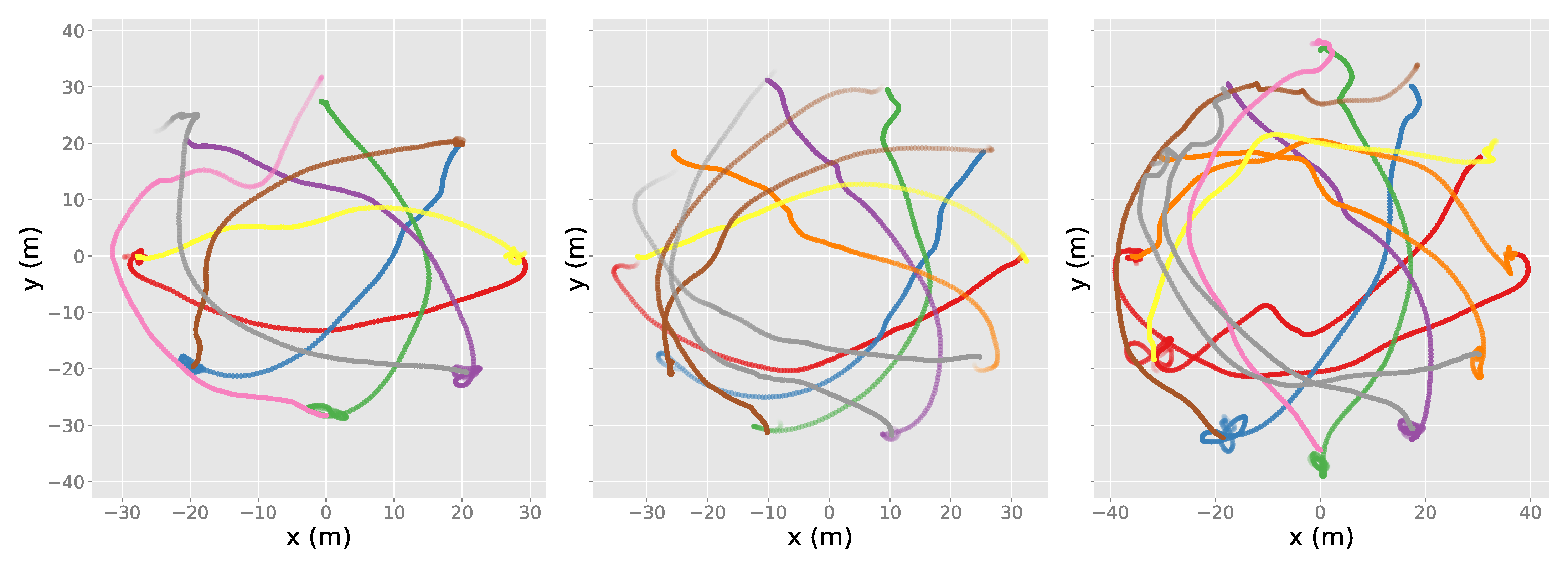}
	\caption{Simulated trajectories of agents for $1$-nearest neighbor BICARL with 2D dynamics. Left: 8 agents; Middle: 10 agents; Right: 12 agents.}
	\label{fig:traj81012}
\end{figure}

For all of our experiments, we use the parameters and hyperparameters indicated in Table~\ref{table:hyperparameters}. We apply curriculum learning \cite{bengio2009curriculum} to improve learning performance, increasing the number of agents continuously while training. The single resulting policy is then evaluated for a varying number of agents to gauge generality and scalability. We consider a collision to occur when agents are closer than $\SI{1.5}{\metre}$ and accordingly tuned the penalty radius during training to $C=\SI{7}{\metre}$ to achieve reasonable behavior for up to $50$ agents.

\subsection{Simulation Results}
We have trained on two commodity desktop computers  equipped with an Intel Core i7-8700K CPU, 16 GB RAM and no GPUs. We successfully and consistently learned good policies in 30 minutes upon reaching e.g. $1.4 \cdot 10^{6}$ time steps. The reason for the fast training lies in the parallelization of collecting experience in Unity, as we can run multiple simulations at the same time to collect data, and the simulation in Unity is fast. We first compare our results to the algorithms FMP \cite{semnani2019forcebased} and ORCA \cite{alonso2013optimal}. As mentioned earlier, for comparison we may also consider our dynamics as part of the algorithm for simpler single integrator dynamics.

\paragraph{Learned collision avoidance behavior}
In the package delivery scenario, we place the agents on a circle of fixed radius and gradually increase the number of agents. As soon as agents become closer than $\SI{3.5}{\metre}$ to the goal, we sample a new goal on the circle. In Table~\ref{table:circle-comparison}, it can be observed that the average packages collected per drone decrease as the number of drones increases, as the drones will increasingly be in each other's way. Further, for the package delivery task, FMP and ORCA eventually run into a deadlock. An example for FMP can be seen in Fig.~\ref{fig:fmp_stuck_circle}. In this case, our methodology provides robust, collision-free behavior that avoids getting stuck.

\begin{figure}
	\centering
	\includegraphics[width=0.95\linewidth]{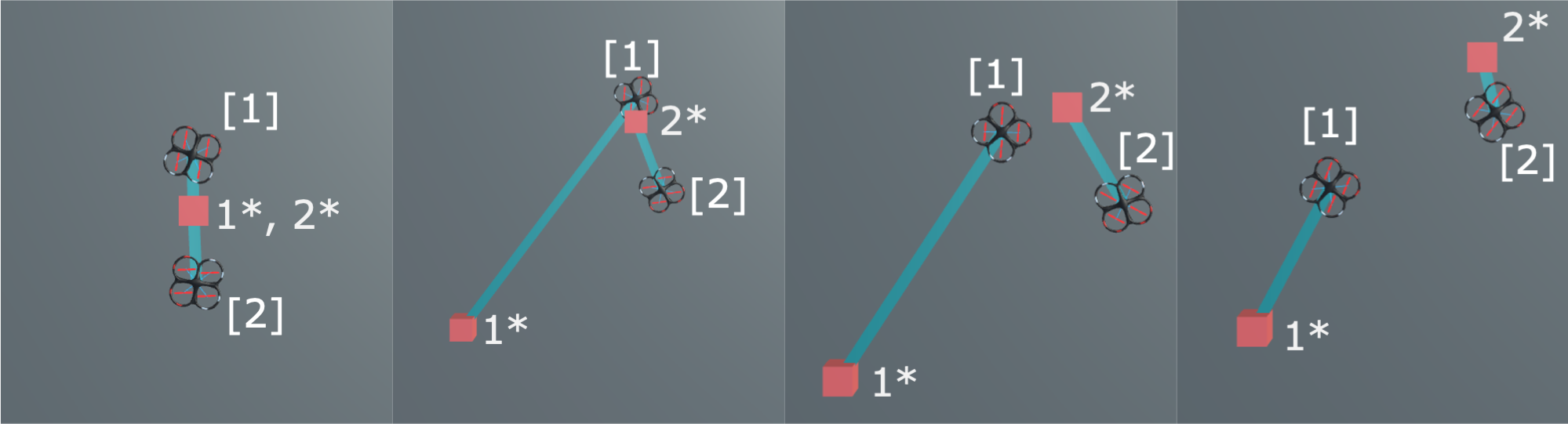}
	\caption{Simulated flight behavior of two agents in the package delivery scenario with overlapping goals $(1^*, 2^*)$. In our approach, one drone yields and the other drone collects, obtaining a new goal. In FMP, the drones become stuck, while in BICARL one drone yields for the other drone. The red squares and blue lines have the same meaning as in Fig.~\ref{fig:fmp_stuck_circle}. Time progresses from left to right.}
	\label{fig:same_target}
\end{figure}

In the formation change scenario, during training we start with 4 diametrically opposed agents on a circle of radius $\SI{70}{\metre}$ and gradually increase the size of the swarm up to 40. It can be seen in Fig.~\ref{fig:traj81012} that rotation around a fixed direction emerges. Increasing the number of agents leads to situations where ORCA and FMP get stuck. \cite{trautman2010unfreezing} demonstrates that the solution is cooperative collision avoidance. In line with this finding, our learned policy is able to capture such cooperation, i.e. one drone gives way to another as can be seen in Fig.~\ref{fig:same_target} and supplementary videos. 

Overall, from Table~\ref{table:comparison} we see that our solution is competitive, especially for many agents. Note that extra time to goal and travelled distance are measured relative to the baseline where drones fly in straight lines to the goal. Although FMP achieves very good results for small numbers of agents, at more than 35 agents FMP becomes stuck, while our method learns a robust behavior that works for large numbers of agents. Here, we tuned FMP parameters to obtain reasonably smooth flight. While improved results in FMP could be possible, additional parameter tuning would be required. 

\begin{figure}
	\centering
	\includegraphics[width=0.95\linewidth]{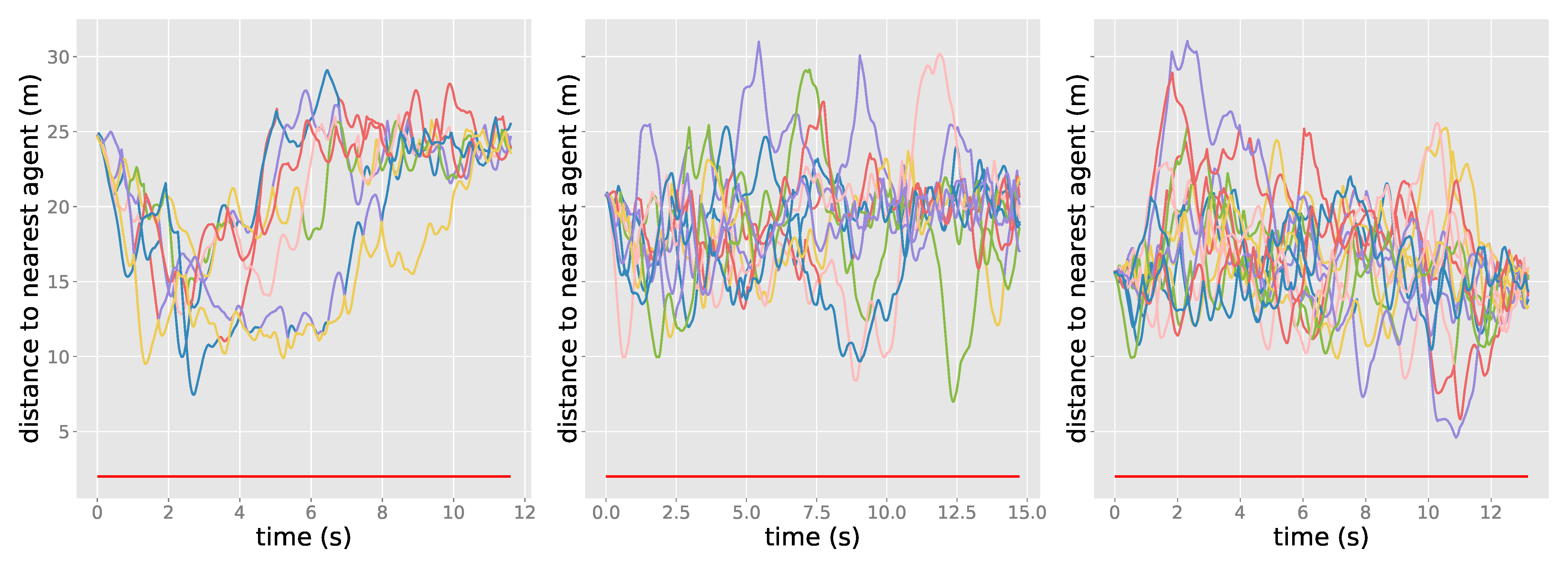}
	\caption{Simulated minimum inter-agent distances achieved in the circle formation change scenario. The red line indicates the radius considered as collision. Left: 10 agents; Middle: 15 agents; Right: 20 agents.}
	\label{fig:dis_101520}
\end{figure}

In Fig.~\ref{fig:dis_101520}, we find that our method successfully avoids collisions with other agents while reaching the goals. During training, collisions may be caused by nearby other agents regardless of the own behavior. As a result, this often provides a negative feedback signal even if the drone itself is not responsible for the collision, resulting in behavior where agents avoid other agents even when the punishment for violating the avoidance radius (here $C=7$) is far off. 

\begin{figure}
	\centering
	\includegraphics[width=0.95\linewidth]{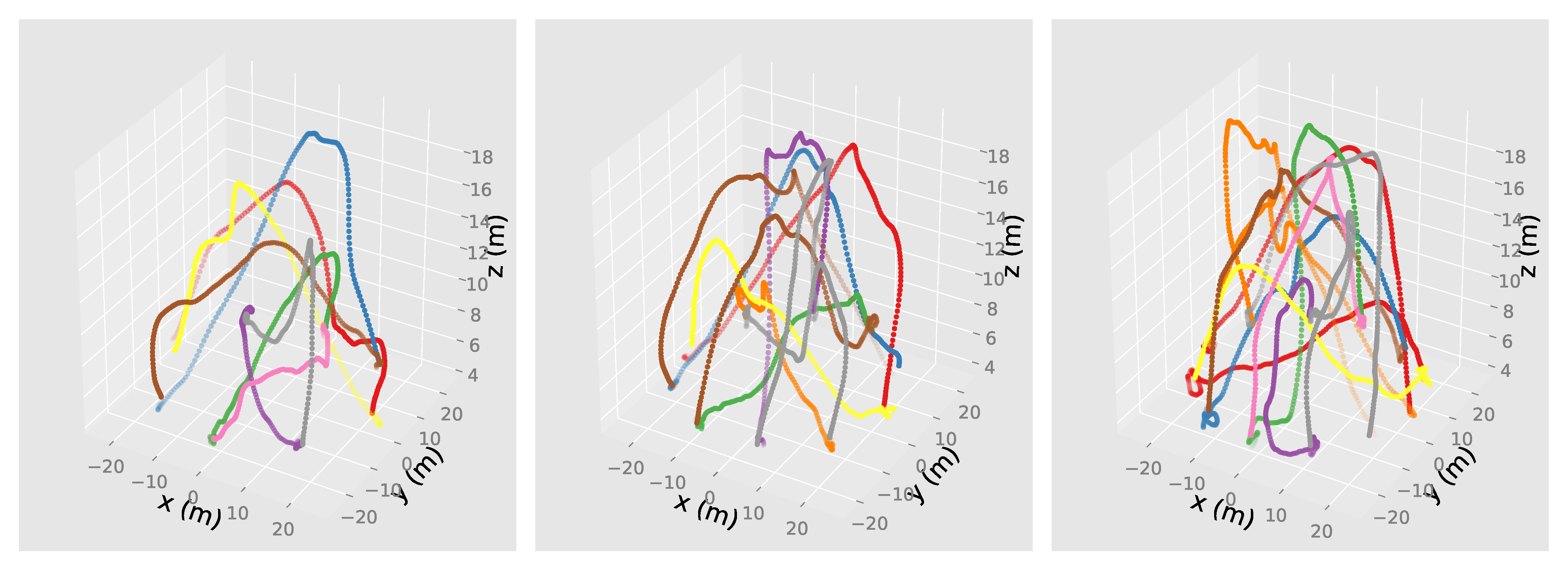}
	\caption{Simulated trajectories of agents for $1$-nearest neighbor BICARL with 3D dynamics. Agents have free space above their starting position and use it to navigate past each other. Left: 8 agents; Middle: 10 agents; Right: 12 agents.}
	\label{fig:traj3D_81012}
\end{figure}

Analogously to the 2D case, we can consider formation change in the 3D case. In Fig.~\ref{fig:traj3D_81012} the trajectories of our learned policy are depicted. It can be seen that our methodology remains successful in guiding the agents to their goal destination. Furthermore, to make use of the additional space above, agents begin flying above and below each other.

\begin{figure}
	\centering
	\includegraphics[width=0.95\linewidth]{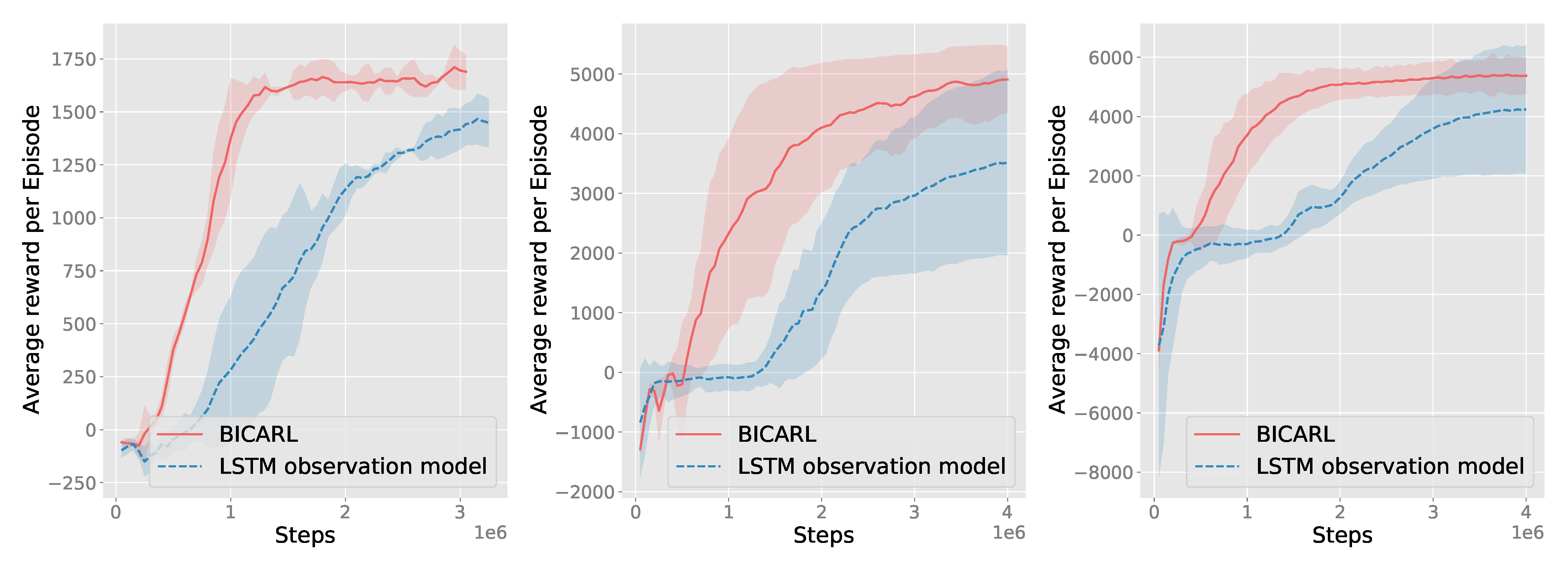}
	\caption{Learning curve: Average achieved sum of rewards per episode over total time steps, plus variance as shaded region for three random seeds. It can be seen that our model learns significantly faster. Left: 2 agents; Middle: 4 agents; Right: 6 agents.}
	\label{fig:lstm_bicarl}
\end{figure}

\begin{figure*}
	\centering
 	\includegraphics[width=\linewidth]{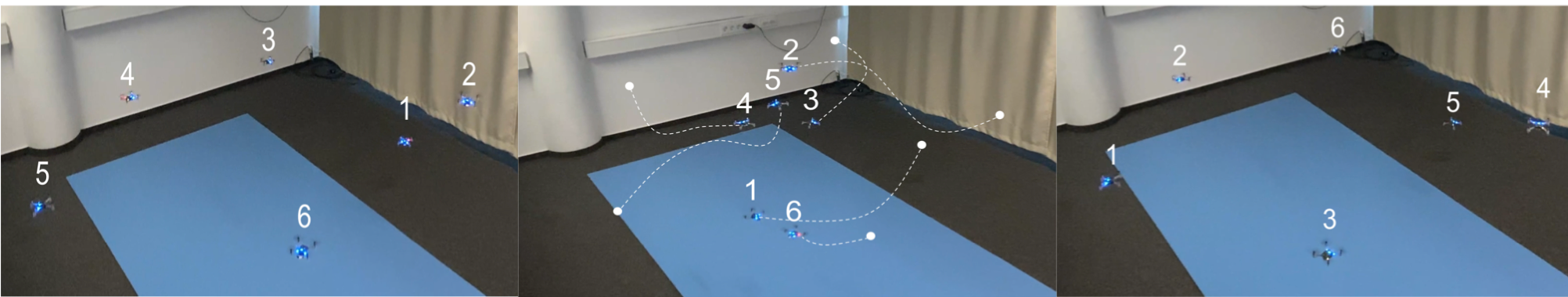}
	\caption{Real world circle formation exchange with the policy for the 3D case embedded on a swarm of 6 Crazyflie nano-quadcopters. The drones successfully switch their positions to the antipodal points on a circle by engaging in 3D collision avoidance. Time progresses from left to right.}
	\label{fig:real}
\end{figure*}

\paragraph{Comparison to LSTM-based policies}
We compare our observation model with the LSTM variant proposed in \cite{everett2021collision}. We use an LSTM observation filter in conjunction with the two-layer architecture and run several tests with different parameters (e.g. number of hidden units) for 2, 4 and 6 agents, comparing the average achieved reward in each case.

The learning curves in Fig.~\ref{fig:lstm_bicarl} suggest that our observation model achieves comparable results and faster learning. As expected, training LSTMs is slow and our method speeds up learning greatly, especially for many agents. While this does not exclude that in general and for improved hyperparameter configuration, the LSTM observation model will be superior, our method regardless has less parameters to fine-tune, has a constant computation cost regardless of the number of agents and hence is cheaper to both train and apply in the real world. 

Although in \cite{everett2021collision} it has been noted that learned collision avoidance algorithms such as \cite{chen2017socially} based on neighborhood information tend to get stuck in large swarms or cause collisions, we obtain a seemingly opposed result. The reason is that our double integrator model together with the added yaw angle allows agents to avoid getting stuck by implicitly saving information or communicating with each other via the yaw angle. Since the problem at hand is multi-agent, a general optimal solution should act depending on the whole history of observations and actions. If two drones are stuck, they can therefore remember past information and change their reaction, leading to better results. Indeed, in our experiments we found that the yaw angle is crucial to obtaining good performance, leading us to the conclusion that a sufficiently rich motion model may in fact ease the learning of good collision avoidance behavior.

\subsection{Real World Experiments}
We validate the real-world applicability of our policies by performing experiments on swarms of micro UAVs. We downscale lengths and speeds by $10$ (since the used indoor drones are only $\SI{0.1}{\metre}$ in length). We directly apply the desired velocity as a set point for the low-level controller instead of the acceleration as in conventional double integrator models, which is not easily regulated due to the non-linearity of typical quadrotor dynamics.

\paragraph{Hardware Setup} 
For our real-world experiments, we use a fleet of Crazyflie nano-quadcopters \cite{giernacki2017crazyflie} with indoor positioning via the Lighthouse infrared positioning system and extended Kalman filter \cite{mueller2015fusing} with covariance correction \cite{mueller2017covariance}. Since the Crazyflies cannot sense each other, we simulate the inter-drone observations by exchanging information via TDMA-based peer-to-peer radio communication to broadcast the estimated position and velocity of each drone. Although the sensor range $K$ in the simulation can model a limited range of wireless communication, in our real indoor experiments there is no such limitation. Instead, drones ignore all information other than their supposed observations. As low-level controller, we use the default cascaded PID controller of the Crazyflie, setting desired velocities and yaw rates from the output of our policy directly as set point of the controller. Note that our code runs completely on-board, except for starting and stopping experiments. 

\paragraph{Evaluation}
We find that the policies obtained in simulation work as expected for two to six drones in reality, and expect similar results for more real drones. In Fig.~\ref{fig:real}, it can be seen that the drones successfully complete formation changes. Using our modified double integrator model, we easily achieve stable results in real world tests, as the cascaded PID low-level-controller is able to achieve desired velocities while stabilizing the system. Due to the simulation-to-reality gap stemming from model simplifications, we can observe slight control artefacts (oscillations). Nonetheless, the combination of traditional controller with learned policy successfully realizes the desired collision avoidance behavior, and the on-board inference time of the policy remains below $\SI{1}{\milli \second}$, with $\SI{30}{\micro \second}$ on our PCs. As a result, by using our motion model of medium complexity, we obtain a practical approach to decentralized quadrotor collision avoidance.
 
\section{CONCLUSION}
To summarize, in this work we have demonstrated that information of $k$-nearest neighbors, non-recurrent policies and a sufficiently rich motion model are enough to find robust collision avoidance behavior. We have designed a high-level model that allows application of the learned policy directly to the real world in conjunction with e.g. a standard cascaded PID controller. In the large swarm case and for collecting packages, our scalable and decentralized methodology appears to show competitive performance. Furthermore, our method avoids getting stuck and is very fast to learn. 

Interesting future work could be to consider external static or dynamic obstacles such as walls. One could combine our approach for $k>1$ with Deep Sets \cite{zaheer2017deep}, mean embeddings \cite{huttenrauch2019deep} or attention mechanisms \cite{manchin2019reinforcement} to further improve learning behavior. Finally, it may be of interest to further reduce the simulation-to-reality gap by using a more detailed model that models e.g. a 12-state quadrotor dynamics with simulated PID controller, motor lag, detailed drag models etc.



\bibliographystyle{IEEEtran}
\bibliography{IEEEabrv,lit}

\end{document}